\documentclass[letterpaper]{article}
\usepackage{aaai} 
\usepackage{times}
\usepackage{helvet}
\usepackage{courier}
\usepackage{url}
\usepackage{graphicx}
\usepackage{epstopdf}
\usepackage{pgfplots}
\usepackage{tikz}
\usetikzlibrary{shapes.geometric, arrows}
\usepackage{textcomp}
\usepackage{tabu}
\usepackage{wrapfig}
\usepackage[font=small,labelfont=bf]{caption}
\newcommand{\comment}[1]{}
\begin{document}
\pdfinfo{ /Title ("How Was Your Weekend?" A Generative Model of Phatic Conversation) /Author (Hannah Morrison, Chris Martens) /Keywords (generative model, communication, conversation modeling, AI) } %
\title{``How Was Your Weekend?''\\A Generative Model of Phatic Conversation}

\author{Hannah Morrison \and Chris Martens\\\{hmorris3, crmarten\}@ncsu.edu\\Principles of Expressive Machines Lab\\
Department of Computer Science\\
North Carolina State University\\
Raleigh, NC
} 

\maketitle

\begin{abstract}
Unspoken social rules, such as those that govern choosing a proper discussion topic and when to change discussion topics, guide conversational behaviors. We propose a computational model of conversation that can follow or break such rules, with participant agents that respond accordingly. Additionally, we demonstrate an application of the model: the Experimental Social Tutor (EST), a first step toward a social skills training tool that generates human-readable conversation and a conversational guideline at each point in the dialogue. Finally, we discuss the design and results of a pilot study evaluating the EST. Results show that our model is capable of producing conversations that follow social norms.
\end{abstract}

\section{Introduction}

Social interactions are complex, each regulated by a set of social norms. In successful interactions, these norms are followed. However, these norms vary. What might be considered a violation of a social norm in one situation may be acceptable, and even expected, in another. Context, which determines the type of interaction, matters~~\cite{rapport}. Social interaction is further complicated due to its reliance on social cues: facial expressions, body language, and other nonverbal behaviors contribute information to conversations~\cite{nonverbal}.

We are interested in building computational models of social-norm-following dialogue so that we can generate conversations and use them in order to simulate believable interactions between virtual agents. Given the inconsistency of conversational norms, we focused our attention on a common, yet complex, interaction: a short encounter between two acquaintances. We created a rule-based, generative model that simulates conversation between two agents, who may follow and violate social norms and generate emotional responses to the conversation.

This work contributes to the landscape of AI tools available for social skills training. We present a prototype social skills training tool (SSTT) as an example application of our model, called the Experimental Social Tutor (EST).  We designed and ran a pilot study in order to evaluate the EST's usability and quality of conversations generated by the model. Results suggest that although the rendering of natural language between characters needs improvement, the EST's model is able to generate conversations that follow social norms closely.   

\section{Background: Defining Conversational Rules}
We seek to model interactions between two acquaintances. We base our model on the following rules, taken primarily from axioms of interpersonal interaction presented by Berger and Calabrese~\cite{initial}, which represent social norms and attitudes:
the goal of casual conversation among acquaintances is to improve rapport; following the social norms of greeting and exchanging pleasantries is favored because conversational participants know what to expect; participants ask more questions during the beginning of conversation, and offer more detail as conversation continues;participants value self-disclosure; participants favor mutual participation; participants view their conversational partner more favorably if they have similar interests or opinions (thus, participants should seek to find commonalities).

Berger and Calabrese state that the goal of conversation between two strangers is to reduce the amount of uncertainty between them---this is why unexpected conversational moves (i.e. social gaffes), such as dominating a conversation, make conversation less likely to continue. Because commonalities between conversational participants reduce uncertainty, participants prefer speaking with those that are similar to them. The importance of similarity among participants is echoed by Byrne and Griffitt~\cite{interpersonal}, who conclude that attraction can be estimated by the number of similar opinions between the participants. They found that this holds among children, hospital patients, and the elderly: that is, that  the importance of similarity between conversational participants is ubiquitous. 

Similarity can be found through the establishment of the common ground between participants. Cassell et al.~\cite{sociallanguage} state that familiarity among participants can be increased by talking about ``topics that are obviously in the common ground such as the weather, physical surroundings, and other topics available in the immediate context of utterance." Participants may also increase familiarity by introducing personal information, thereby moving it into the common ground. Zegarac~\cite{phatic} states that the goal of phatic (small-talk-like) communication is to manage the common ground between participants. 

We use the sources above to model socially normative conversation as conversation that engenders rapport among participants. Conversation that violates social norms, then, would decrease rapport. A deviation from the rules we present above would constitute this. For the scope of our task, however, we limit the social violations in the model to the following: participants that do not involve the other participant (that is, those that dominate their conversation), or participants that vocalize a strong disagreement in opinion, will decrease the chance that the conversation will continue, and lower rapport between them and their conversational partner.  
\section{Related Work} 
PsychSim, created by Marsella et al.~\cite{psychsim}, is a “social simulation tool” that can be used to generate a variety of social situations. Agents in the simulation may communicate with each other and each have their own goals, opinions, and beliefs regarding others. The purpose of this tool is for the user to explore social situations in a casual, low-pressure environment.  While the tool we present in this paper is much simpler---the model's agents do not hold beliefs about others beyond whether or not they are liked, nor do they have explicitly encoded goals---we see our simple, logic-based encoding as having several advantages, such as ease of authoring and extension, the ability to trace causal links between utterances, and the separability of the model from its specific implementation in a SSTT.

Bickmore and Cassell~\cite{rea} describe an intelligent virtual agent, REA, capable of engaging in small-talk with a user. The agent's conversational choices, and the underlying logic that motivates them, take into account the common ground between participants, the norm of staying on-topic in a conversation, and trust between participants. Our model uses similar social norms, but where REA uses small talk in the context of selling real estate, our model is general over topics and may be used to generate conversations for many settings.

Previous work by Even et al.~\cite{sSSTT} introduced an interactive, turn-based SSTT for schizophrenia patients that can ``teach the patients how to introduce themselves, start, maintain or interrupt a discussion, request or refuse something, receive or give a compliment, or criticism.'' Tartaro and Cassell~\cite{peersSSTT} describe a virtual peer with which the user can interact. Users engage in a `collaborative storytelling task" with a life-sized avatar and may also manipulate the nonverbal behaviors of the virtual peer in order to ``observe the effects on interaction.'' Our proposed SSTT differs from these in several ways, most notably in its focus on the casual relationships between conversational events and our more explicit presentation of the social rules we aim to teach. While the EST may have the potential to be useful on its own, we envision the EST in its current state as a complement to these avatar-containing SSTTs, being used, perhaps, prior to these in order to ``warm up" participants for conversation with the virtual avatar.

\section{The Model}
\begin{figure}
\centering
\includegraphics[scale=.16]{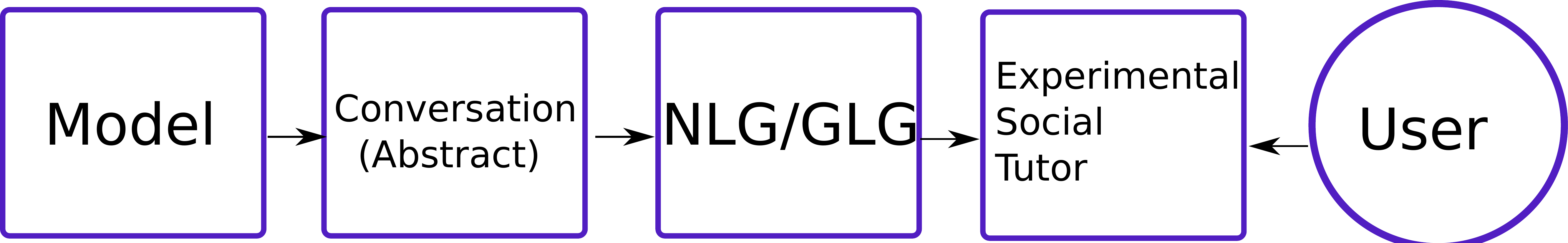}
\caption{A diagram of the project's architecture. The model produces an abstract conversation, which is then converted into text by a natural language generator (NLG) and guideline generator (GLG).}
\label{fig:estarch}
\vspace{-1.5em}
\end{figure}
We take inspiration from several existing models of social interaction. Richard Evans~\cite{constitutive}, in presenting a model based on the Game of Giving and Asking for Reasons, describes agents that follow the conversational norm of turn-taking and are able to express emotions such as worry. Dixon et al.~\cite{PAD} present a framework for modeling complex agents that introduces the concept of dialogue-related obligations, encoding the social norms of requesting and receiving information in an exchange between two individuals.

Our executable model generates conversations that may either follow or break social norms, with participant agents that respond accordingly. When the model is run, it non-deterministically generates one of many distinct conversations between two acquaintance agents. Agents must follow the norms of greeting one another and saying goodbye before exiting the conversation, must stay on topic, and must make small talk in some form before moving onto more varied topic discussion\footnote{Initial topics are restricted to the \emph{most} common ground (this idea is taken from \cite{sociallanguage}): the agent's weekend or the weather; additional topics are sports (baseball, soccer, or running) and music (pop, country, or rock).}. Agents are, however, free to offend, bore, or annoy their conversation partner, and the partner, if made sufficiently upset, may leave the conversation. In order to simulate a time constraint, each utterance spends a turn. Conversation will transition to a natural ending after a set number of turns.

We formalize these conversational moves and conventions as an unordered collection of rules describing what is possible for the agents to do under which circumstances. Conversation states are encoded as collections of logical predicates: for example, when Alice becomes annoyed, the predicate \texttt{feels(alice, annoyed)} will be added to the conversational state.  

\tikzstyle{rule} = [rectangle, rounded corners, minimum height=.5cm,text centered, draw=black, fill=blue!10]
\begin{figure}[h]
\centering
\includegraphics[scale=.20]{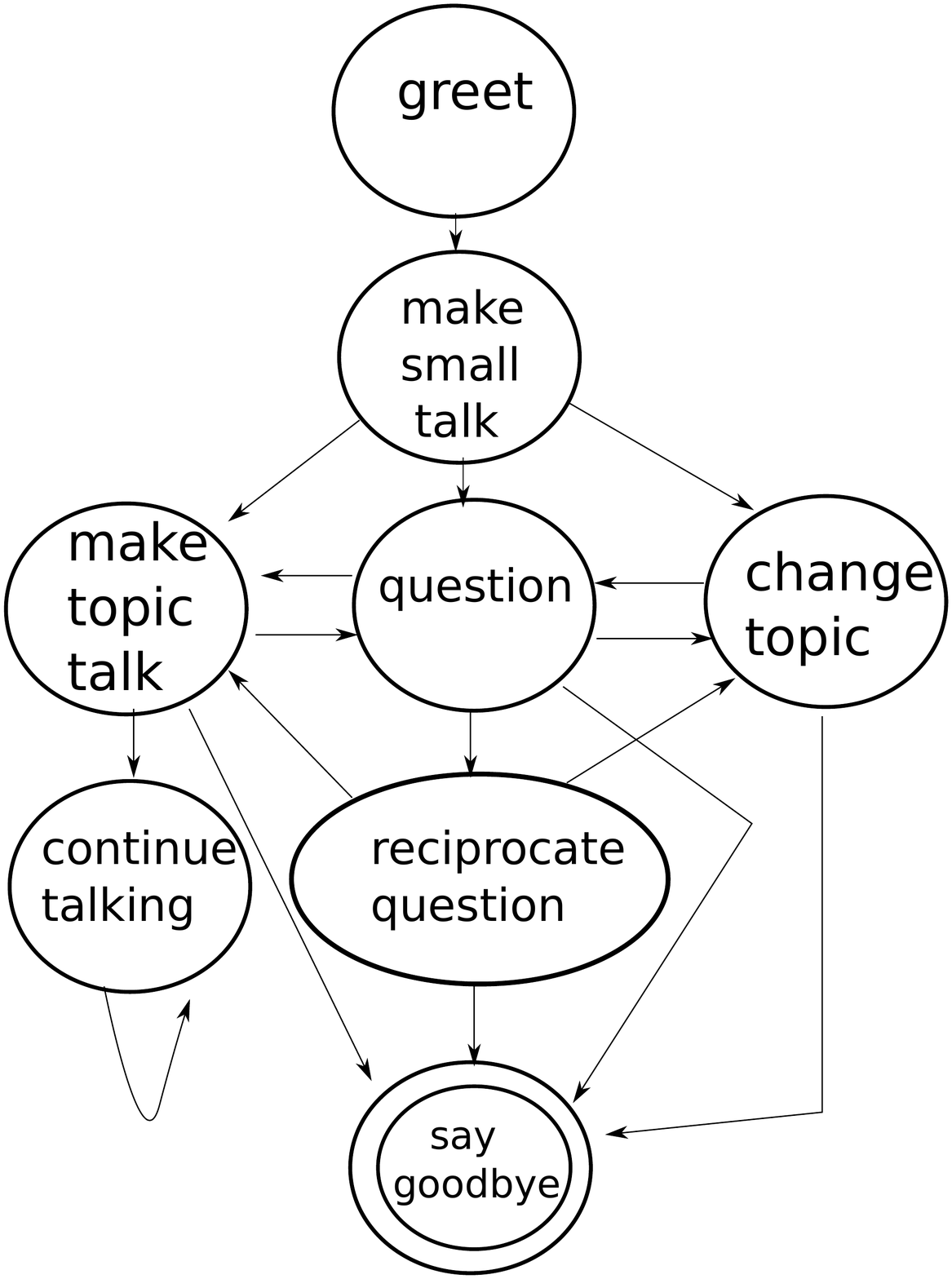}
\caption{A finite state machine depicting the conversation flow constraints in the model. Each state represents a conversational rule (edges between the \textbf{make topic talk} state and \textbf{question} and \textbf{change topic} states have been omitted. Edges between the \textbf{continue talking} state and \textbf{question} and \textbf{change topic} states have also been omitted.) A socially inappropriate conversation might consist of the path {\bf greet$\to$make small talk$\to$make topic talk$\to$continue talking$\to$ continue talking$\to$ continue talking$\to$continue talking$\to$ goodbye}, while a socially normative conversation might consist of the path {\bf greet$\to$make small talk$\to$make topic talk$\to$question$\to$reciprocate question$\to$change topic$\to$topic talk$\to$question$\to$goodbye}.}
\hrule
\label{fig:model}
\vspace{-1 em}
\end{figure}

We represent the following conversational states that may change as conversation proceeds:
\begin{itemize}
\setlength\itemsep{0em}
\item The current topic
\item Each agent's current feeling (happy, sad, annoyed, or content)
\item The number of times each agent has spoken so far
\item Affinity between agents (how much they like each other)
\end{itemize}

In addition, we represent agents having opinions (positive or negative) about topics, and relatedness between topics. These may influence conversation, but are not themselves altered as conversation proceeds.

\subsection{Conversational Moves}

Ryan et al.~\cite{dialoguemanager}  present a procedural system of dialogue generation in which one line of dialogue is generated per conversational turn; each line of dialogue has a corresponding \emph{dialogue move} or moves in the system. The EST works in a similar manner: each line of displayed dialogue corresponds to a rule in the model that has fired.
The model defines dialogue moves for greeting, making small talk, making more advanced 
conversation (``topic talk''), continuing to speak on the current topic, asking a question, reciprocating questions (e.g. ``What about you?"), responding with a typical level of enthusiasm and a high level of enthusiasm (e.g. ``I like soccer" vs ``I love soccer"), changing a topic, and saying goodbye. 
These may, in combination, produce conversation that is either socially appropriate or inappropriate; in either case, agents will respond emotionally. There are no restrictions on the social appropriateness of the generated conversations. 

Conversation is constrained to follow a particular global structure so that, for example, participants may not continue talking before responding to a question, or change the subject without it being related to the current topic. We depict the global conversation structure as a state machine in Figure~\ref{fig:model}, which approximates these constraints as precedence ordering between conversational moves. 

\begin{figure}[h]
\small

\begin{tabu} to 0.5\textwidth { | p{3.5cm} | p{2cm} | p{2cm} | }
 \hline
 \textbf{Interpreted Utterance} & \textbf{Ceptre Rule} & \textbf{State Change} \\
 \hline
\textbf{Bob:} Good morning, Alice!  & greet\_bob\_alice   & greet bob alice  \\
\hline
\textbf{Alice:} Good morning, Bob!  & greet\_alice\_bob & greet alice bob \\
\hline
\textbf{Bob:} This weather today is really nice--good for playing sports  &small\_talk\_
weather & current topic: weather, times spoken by bob: 1\\
\hline
\textbf{Bob:} I did a lot of playing baseball on Saturday It was nice out, just like today.  & change\_topic\_
weather\_
baseball & current topic: baseball, times spoken by bob: 2 \\
\hline
\textbf{Bob:} I think baseball is a lot more interesting than people give it credit for. &  topic\_talk\_
baseball\_
typical\_
positive & times spoken by bob: 3\\
\hline
\textbf{Bob:} Some of the people I know like baseball. & continue\_
talking\_baseball &  times spoken by bob: 6, Alice feels annoyed \\
\hline
\textbf{Alice:} Uh-huh, well...I have to go now. Goodbye. & annoyed\_by\_
unfair\_
participation, 
terminate\_
conversation & Alice feels annoyed, conversation is ending \\
\hline
\textbf{Bob:} Take care. & say\_goodbye\_
bob\_alice
& say\_goodbye\_
bob\_alice \\
\hline
\end{tabu}

\caption{An example of an abbreviated model-generated sequence of statements. Some of Bob's statements have been omitted.}
\label{fig:example}
\vspace{-1.8em}
\end{figure}

Here are two examples of the conversational moves we implement:
\begin{itemize}
\setlength\itemsep{0.01em}
\item {\bf Make topic talk:} If the current topic is $T$ and agent $C$ has an opinion $O$ about $T$,
			then $C$ states that their opinion about $T$ is $O$ (in either a typical or enthusiastic way).
            Increment the number of times $C$ has spoken.
\item {\bf Reciprocate question:} If $C$ has just asked $C'$ a question about topic $T$, $C$ states their opinion on $T$ and asks the same question in response. Increase affinity between $C$ and $C'$ and increment the number of times $C'$ has spoken.
\end{itemize}

\subsection{Emotional and Affinity Changes.}
Rules that can change agent emotions and affinity between agents include:
\begin{itemize}
\setlength\itemsep{0.01em}
\item {\bf Like from agreement:} If $C$ has just stated their opinion on topic $T$, and $C'$ shares this opinion, increase affinity of $C'$ towards $C$.
\item {\bf Annoyance from unbalanced participation:} If $C$ is feeling content, but the number of times $C'$ has spoken is more than $2/3$ the total conversation length, change $C$'s feeling to annoyed.
\end{itemize}

\subsection{Implementation}
We implemented the model using the linear-logic-based modeling language Ceptre~\cite{ceptre}, which allows for a declarative representation of conversational moves. Each move is represented with a name, a set of preconditions, and a description of how it modifies the conversational state. The set of moves is unordered, but ordering constraints may be introduced through preconditions, e.g. to enforce the structure described in Figure~\ref{fig:model}. Once the author provides an initial state (a set of facts in first-order logic), Ceptre will run the state forward according to the moves that apply, choosing nondeterministically whenever multiple moves apply, and stop when no more moves apply. The nondeterminism inherent in Ceptre can generate highly varied conversational output.
Figure~\ref{fig:example} shows an example conversation generated by the model.

\section{The Experimental Social Tutor}
Socialization is sometimes problematic for people on the autism spectrum. Autism is a neurodevelopmental condition characterized by difficulty interacting and communicating with others (e.g. having trouble interpreting facial expressions) ~\cite{peersSSTT} as well as restricted, repetitive behavior (e.g. becoming anxious if one's routine is not followed) and interests (e.g. having an intense focus on a special interest or topic) ~\cite{repetitive}.  While those on the spectrum can often identify nonverbal and social cues in isolation, they may have difficulty interpreting them in practice, which can lead to problems such as loneliness, social isolation, and avoidance of social interaction~\cite{children}. We are interested in developing computational tools that may be useful in helping people on the spectrum learn about social interactions. The EST is an early prototype of such a tool.

The EST separates the concerns of structure and presentation of conversation. In addition to our rule-based {\em model} of social interaction described above, which generates the abstract structure of conversation in terms of conversational moves, we implemented a front-end {\em interface} to render and contextualize the conversations.
The front end transforms the abstract representation into concrete dialogue and an accompanying social guideline. This text is displayed to the user by the EST, our early-stage, web-based SSTT. The architecture of our system is depicted in Figure~\ref{fig:estarch}.

\subsection{Dialogue and Guideline Generation}
The EST uses conversations generated by the model described above to display a natural-language rendering of the conversation and relevant conversational guidelines. It is intended for middle-and-high-school aged people on the autism spectrum. Figure 4 displays the prompt given to the user when beginning use of the application. Figure 5 displays sample output of the EST.\\
\begin{figure}
\centering
\fbox{
		\includegraphics[scale=.4]{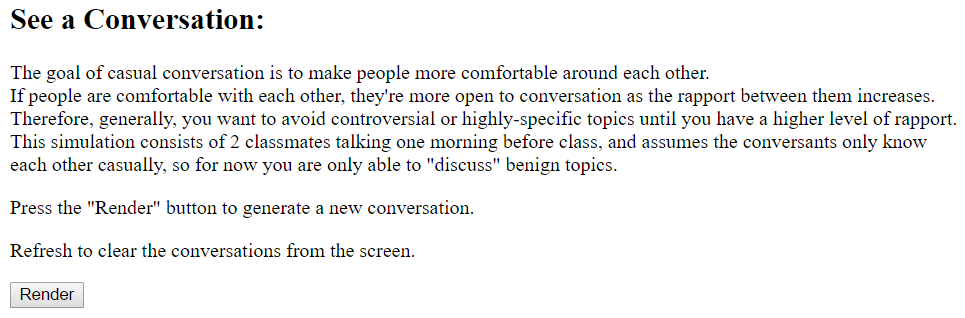}
}
	\caption{A partial screenshot of the EST}
\vspace{-1.3em}
\end{figure}
\begin{figure}[h]
\centering
\begin{tabu} to 0.5\textwidth { | X[l] | X[c] | }
 \hline
 \textbf{Dialogue} & \textbf{Guideline}  \\
 \hline
Bob: Good morning, Alice.& Good places to start a conversation: waiting in line, a club  meeting, on the bus. \\
\hline
Alice: Good morning, Bob & Greeting someone acknowledges them and lets them know you are open to conversation. \\
\hline
Alice: I love time at home listening to my music. Wish the weekend didn't go by so quickly & Small talk makes people more comfortable around each other. \\
\hline
Bob: How was your weekend, Alice? & Avoid only asking questions and never giving information about yourself; this makes the conversation one-sided. \\
\hline
\end{tabu}
\caption{A table depicting a partial example of EST output.}
\vspace{-1.5em}
\end{figure}

\subsubsection{Dialogue Generation}

Our dialogue generator takes as input a trace from Ceptre abstractly representing a conversational exchange, and produces human-readable text output for that conversation. It maps each step of the trace to a line of dialogue, where the rule in the model that determined the step determines an utterance to represent it.
In order to increase diversity of phrasing, some cases have multiple potential utterances, which are displayed at random.

\subsubsection{Guideline Generation}
As with agent dialogue, each guideline is generated by case-analyzing the rule used in the model. In order to communicate numerous relevant ideas, some cases have multiple potential guidelines, which are selected to appear at random. 

Guidelines were gleaned from several sources.  Our primary reference was a social-skills book written by an autistic adult~\cite{social}. We also referenced several guides describing therapeutic sessions for social-skills improvement~\cite{PEERS,children} and a guidebook intended for autistic teenagers~\cite{day}. Rules in the model that involved the initial stages of conversation were matched with guidelines about when and where to initiate conversation, as well as the importance of beginning conversation; rules involving subsequent dialogue moves were matched with guidelines about the importance of making one's conversational partner comfortable through near-equal exchange of information, as well as potential conversation topics; rules involving questions were matched with guidelines explaining the role of questions as signals of interest, in addition to advising how often questions should be asked.

\section{Experiment}
Next, we describe the design and results of a pilot study evaluating the EST. 
\subsection{Design}

Because the EST is in an early stage of development, our primary goal in study design was to evaluate the model's capability of generating conversation that follows social norms. 
The questions asked included three 0-10 scales asking participants to judge difficulty of using the EST, realism of the conversations, and how closely the conversations follow social norms.

Due to the early stage of development of the EST, we did not test the EST's intended users: middle-and-high schoolers on the spectrum. However, we were able to get feedback from 3 autistic college-aged participants. 
Participants (N=35) were recruited through advertising (flyers, in-class announcements, etc.) on campus. The study was conducted entirely online: participants were e-mailed instructions that included a link to the tool and a link to the survey (which, when completed, re-directed to another survey where participants could enter their e-mails for compensation purposes, in order to ensure participant anonymity). Participants were e-mailed \$10 Amazon gift-cards for their participation.

\subsection{Results}
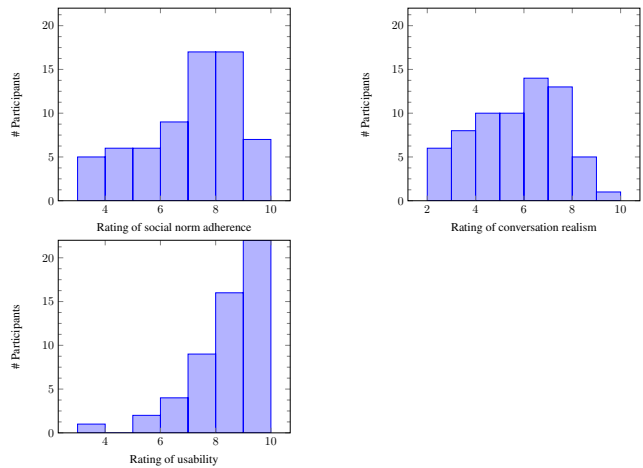
\begin{figure}[h]
\begin{tikzpicture}[scale=0.45]
\begin{axis}[
    ymin=0, ymax=22,
    minor y tick num = 3,
    area style,
    ylabel= \# Participants,
    xlabel=Rating of social norm adherence
    ]
\addplot+[ybar interval,mark=no] plot coordinates { (3, 5) (4, 6) (5, 6) (6, 9) (7, 17) (8, 17) (9,7) (10, 0) };
\end{axis}
\end{tikzpicture}
\begin{tikzpicture}[scale=0.45]
\begin{axis}[
    ymin=0, ymax=22,
    minor y tick num = 3,
    area style,
    ylabel= \# Participants,
    xlabel=Rating of conversation realism
    ]
\addplot+[ybar interval,mark=no] plot coordinates { (2, 6) (3, 8) (4, 10) (5, 10) (6, 14) (7, 13) (8,5) (9, 1) (10,0) };
\end{axis}
\end{tikzpicture}
\begin{tikzpicture}[scale=0.45]
\begin{axis}[
    ymin=0, ymax=22,
    minor y tick num = 3,
    area style,
    ylabel= \# Participants,
    xlabel=Rating of usability
    ]
\addplot+[ybar interval,mark=no] plot coordinates { (3, 1) (4, 0) (5, 2) (6, 4) (7, 9) (8, 16) (9,22) (10, 0) };
\end{axis}
\end{tikzpicture}

\caption{Histograms depicting the range of participant ratings of social norm adherence, conversation realism, and usability, on a scale from 0 to 10.}
\label{fig:histograms}
\end{figure}

Most participants found the tool easy to use and thought that the generated conversations followed social norms; however, they found the dialogue generated by the tool to be awkward and repetitive. See Figure~\ref{fig:histograms} for histograms of ratings for all three questions asked.

\subsubsection{Social Norm Adherence} Most participants rated the generated conversations as adhering to social norms closely (median: 8). Many mentioned\footnote{For our purposes, `mentioned', `said', etc. refer to the text response that was submitted by the participant.} the one-sided conversation between the two agents, stating ``People will most likely back away from a conversation if it is one sided or confusing" and ``The way they awkwardly end the conversation if the other person is talking too much is really good."

\subsubsection{Realism of Generated Conversation} Most participants found the ``speech" between the two agents to be moderately realistic (median: 6). Some participants felt the speech between the classmates was ``too formal," or ``stiff and unnatural." Despite this, the dialogue still evoked emotional responses from users. One participant said, ``[The conversation] reminds me of a casual conversation I have with someone like a study partner." One participant felt an emotional reaction in response to the conversations: ``Honestly I had some pain from seeing social norms breached by one person. It was like watching a movie where someone messes up really badly." 

\subsubsection{Usefulness}
Interestingly, several neurotypical participants found the EST educational, stating that it prompted them to think about conversation in a way they typically do not. One participant said  ``the `guidline' [sic] section of the tool described things I do in conversation that I didn't even realize I was doing." Another said ``The `guideline' part made me think about how some people...would view social interaction as a set of rules and steps instead of something that happens somewhat naturally and organically."

One autistic participant said that, while the user experience needed significant work, ``the underlying principles are sound." Another said that they had trouble applying social rules rather than identifying them, so ``adding more detail to the guidelines or more specific strategies might be more helpful." (We discuss a response to this in the Future Work section.)\footnote{While we intended for the EST's conversations to act as examples of applications of these rules, they were generally too unrealistic to be helpful in this regard.} Another participant said that the guidelines given by the tool were not useful because ``I have done a lot of studying on [how you are supposed to act during a conversation]." We think the EST might have the potential to be helpful during the studying and learning of basic social rules that the participant mentions, although the tool is too basic for use by most adults. Considering the feedback from the first participant we described, we think the EST might be useful for adults if we add more complex interactions; however, we do think it has the potential to be useful for its intended audience (those of middle-school and high-school ages) with some modifications. Most autistic participants expressed that, due to difficulties with nonverbal communication, a video-based component would have been helpful. This supports our vision of the EST in its current state being used in combination with a SSTT that makes use of a virtual avatar.

\section{Summary}
We have presented a model capable of generating varied, norm-adhering conversation between two acquaintance agents. 
We used a declarative, rule-driven system to encode conversational moves, which supports compositional dialogue generation and the ability to track causal relationships between changes in the conversational state.
The EST is an early-stage social skills training tool that is based on our model. The results of a pilot study suggest the model on which the EST is based generates conversation that follows social norms closely. The model on which the EST is based can easily be extended.

\section{Future Work}
The ultimate goal of future work is to create an interactive environment that uses our model of conversation, allowing the user to both observe casual conversations between simulated agents and engage in interactions with them, while receiving useful conversational guidelines. Such agents might also be useful for increasing the social believability of characters in games~\cite{morrison2017generative}.

We plan to improve the natural-language rendering of the conversations generated by the EST by integrating stylistic variations and studying how these affect perception of the conversation. We would also like to add {\em explanations} for the causal relationships between lines of dialogue and the ``hidden variables'' in the conversational state, such as agent emotions. Finally, we plan to integrate more context and reasoning into the model, such as differentiating between conversations between participants with different relationships (e.g. a peer vs. an authority figure).

\bibliographystyle{aaai}
\bibliography{main}

\begin{thebibliography}{}

\bibitem[\protect\citeauthoryear{Berger and Calabrese}{1975}]{initial}
Berger, C.~R., and Calabrese, R.~J.
\newblock 1975.
\newblock Some explorations in initial interaction and beyond: Toward a
  developmental theory of interpersonal communication.
\newblock {\em Human communication research} 1(2):99--112.

\bibitem[\protect\citeauthoryear{Bickmore and Cassell}{2005}]{rea}
Bickmore, T., and Cassell, J.
\newblock 2005.
\newblock Social dialogue with embodied conversational agents.
\newblock {\em Advances in natural multimodal dialogue systems} 30:23--54.

\bibitem[\protect\citeauthoryear{Byrne and Griffitt}{1973}]{interpersonal}
Byrne, D., and Griffitt, W.
\newblock 1973.
\newblock Interpersonal attraction.
\newblock {\em Annual review of psychology} 24(1):317--336.

\bibitem[\protect\citeauthoryear{Cassell and Bickmore}{2003}]{sociallanguage}
Cassell, J., and Bickmore, T.
\newblock 2003.
\newblock Negotiated collusion: Modeling social language and its relationship
  effects in intelligent agents.
\newblock {\em User Modeling and User-Adapted Interaction} 13(1):89--132.

\bibitem[\protect\citeauthoryear{Dixon, Smaill, and Tsang}{2009}]{PAD}
Dixon, L.; Smaill, A.; and Tsang, T.
\newblock 2009.
\newblock Plans, actions and dialogues using linear logic.
\newblock {\em Journal of Logic, Language and Information} 18(2):251--289.

\bibitem[\protect\citeauthoryear{Evans}{2016}]{constitutive}
Evans, R.~P.
\newblock 2016.
\newblock Computer models of constitutive social practice.
\newblock In {\em Fundamental Issues of Artificial Intelligence}. Springer.
\newblock  389--409.

\bibitem[\protect\citeauthoryear{Even \bgroup et al\mbox.\egroup
  }{2016}]{sSSTT}
Even, C.; Bosser, A.-G.; Ferreira, J.~F.; Buche, C.; St{\'e}phan, F.; Cavazza,
  M.; and Lisetti, C.~L.
\newblock 2016.
\newblock Supporting social skills rehabilitation with virtual storytelling.
\newblock In {\em FLAIRS Conference},  329--334.

\bibitem[\protect\citeauthoryear{Giri}{2009}]{nonverbal}
Giri, V.~N.
\newblock 2009.
\newblock Nonverbal communication theories.
\newblock {\em Encyclopedia of Communication Theory}  690--694.

\bibitem[\protect\citeauthoryear{Lam and Aman}{2007}]{repetitive}
Lam, K.~S., and Aman, M.~G.
\newblock 2007.
\newblock The repetitive behavior scale-revised: independent validation in
  individuals with autism spectrum disorders.
\newblock {\em Journal of autism and developmental disorders} 37(5):855--866.

\bibitem[\protect\citeauthoryear{Laugeson and Frankel}{2011}]{PEERS}
Laugeson, E.~A., and Frankel, F.
\newblock 2011.
\newblock {\em Social skills for teenagers with developmental and autism
  spectrum disorders: The PEERS treatment manual}.
\newblock Routledge.

\bibitem[\protect\citeauthoryear{Marsella, Pynadath, and Read}{2004}]{psychsim}
Marsella, S.~C.; Pynadath, D.~V.; and Read, S.~J.
\newblock 2004.
\newblock Psychsim: Agent-based modeling of social interactions and influence.
\newblock In {\em Proceedings of the international conference on cognitive
  modeling}, volume~36,  243--248.

\bibitem[\protect\citeauthoryear{Martens}{2015}]{ceptre}
Martens, C.
\newblock 2015.
\newblock Ceptre: A language for modeling generative interactive systems.
\newblock In {\em Eleventh Artificial Intelligence and Interactive Digital
  Entertainment Conference}.

\bibitem[\protect\citeauthoryear{Morrison and
  Martens}{2017}]{morrison2017generative}
Morrison, H., and Martens, C.
\newblock 2017.
\newblock A generative model of group conversation.
\newblock In {\em Proceedings of the 12th International Conference on the
  Foundations of Digital Games}, ~66.
\newblock ACM.

\bibitem[\protect\citeauthoryear{Myles}{2003}]{children}
Myles, B.~S.
\newblock 2003.
\newblock {\em Social skills training for children and adolescents with
  Asperger syndrome and social-communication problems}.
\newblock Autism Asperger Publishing Company.

\bibitem[\protect\citeauthoryear{Ogan \bgroup et al\mbox.\egroup
  }{2012}]{rapport}
Ogan, A.; Finkelstein, S.~L.; Walker, E.; Carlson, R.; and Cassell, J.
\newblock 2012.
\newblock Rudeness and rapport: Insults and learning gains in peer tutoring.
\newblock In {\em ITS},  11--21.
\newblock Springer.

\bibitem[\protect\citeauthoryear{Patrick}{2008}]{day}
Patrick, N.~J.
\newblock 2008.
\newblock {\em Social skills for teenagers and adults with Asperger syndrome: a
  practical guide to day-to-day life}.
\newblock Jessica Kingsley Publishers.

\bibitem[\protect\citeauthoryear{Ryan, Mateas, and
  Wardrip-Fruin}{2016}]{dialoguemanager}
Ryan, J.~O.; Mateas, M.; and Wardrip-Fruin, N.
\newblock 2016.
\newblock A lightweight videogame dialogue manager.
\newblock In {\em DiGRA/FDG}.

\bibitem[\protect\citeauthoryear{Tartaro and Cassell}{2007}]{peersSSTT}
Tartaro, A., and Cassell, J.
\newblock 2007.
\newblock Using virtual peer technology as an intervention for children with
  autism.
\newblock {\em Towards universal usability: designing computer interfaces for
  diverse user populations. Chichester: John Wiley} 231:62.

\bibitem[\protect\citeauthoryear{Wendler}{2014}]{social}
Wendler, D.
\newblock 2014.
\newblock {\em Improve your social skills}.

\bibitem[\protect\citeauthoryear{Zegarac}{1998}]{phatic}
Zegarac, V.
\newblock 1998.
\newblock What is" phatic communication"?
\newblock {\em Pragmatics AND Beyond New Series}  327--362.

\end{thebibliography}

\end{document}